\documentclass[journal,twoside,web]{ieeecolor}
\usepackage{amsmath,amssymb,amsfonts}
\usepackage{algorithmic}
\usepackage{algorithm}
\usepackage{array}
\usepackage{cite}
\usepackage[caption=false,font=normalsize,labelfont=sf,textfont=sf]{subfig}
\usepackage{textcomp}
\usepackage{stfloats}
\usepackage{url}
\usepackage{verbatim}
\usepackage{graphicx}
\usepackage{booktabs}
\usepackage{xcolor}
\usepackage{afterpage}
\usepackage{multirow}
\usepackage[caption=false]{subfig}
\usepackage{tikz}
\usepackage[export]{adjustbox}

\usepackage{colortbl}%
\usepackage{arydshln}%
\usepackage{generic}
\usepackage{graphicx}
\usepackage{textcomp}

\makeatletter
\providecommand{\refname}{References}
\makeatother

\def\BibTeX{{\rm B\kern-.05em{\sc i\kern-.025em b}\kern-.08em
    T\kern-.1667em\lower.7ex\hbox{E}\kern-.125emX}}
\begin{document}
\title{Attachment Anchors: A Novel Framework for Laparoscopic Grasping Point Prediction in Colorectal Surgery}
\author{Dennis N. Schneider, Lars Wagner, Daniel Rueckert, \IEEEmembership{Fellow, IEEE}, and Dirk Wilhelm%
\thanks{This work has been funded by the Bavarian Collaborative Research Program (BayVFP)
under der Funding Line Digitization under Grant DIK-2406-0025.}
\thanks{Dennis N. Schneider, Lars Wagner, and Dirk Wilhelm are with the Technical University of Munich, TUM School of Medicine and Health, TUM University Hospital rechts der Isar, Department of Surgery, Research Group MITI, Munich, Germany}%
\thanks{Daniel Rueckert is with the Technical University of Munich, TUM School of Medicine and Health, TUM University Hospital rechts der Isar, Chair for AI in Healthcare and Medicine Munich, Munich, Germany and with the Department of Computing, Imperial College London, London, UK}%
}

\maketitle

\begin{abstract}
Tissue grasping is a fundamental part of surgical tissue manipulation.
Its automation poses a unique opportunity to incorporate into surgical robotics due to its fundamentally non-destructive nature and feasible human oversight.
From a machine-learning perspective, colorectal surgery provides an attractive application setting due to its repetitive tissue manipulation patterns progressing along the tubular colon organ.
In this work, we introduce means to find landmarks in colorectal surgery which suffers from a severe lack of visually distinct feedback.
We introduce \emph{attachment anchors}, a structured representation of the local tissue mechanics to generate robust grasping predictions by normalizing surgical scenes into a consistent local reference frame. We demonstrate that attachment anchors can be predicted from laparoscopic images and incorporated into a grasping framework based on machine learning. Experiments on a dataset of 90 colorectal surgeries demonstrate that attachment anchors improve grasping point localization compared to image-only baselines by \textgreater{10\%}. There are particularly strong gains in out-of-distribution settings, including unseen procedures, operating surgeons and an unseen surgical center. These results suggest that attachment anchors are an effective intermediate representation for learning-based tissue manipulation in colorectal surgery. 
\end{abstract}

\begin{IEEEkeywords}
Geometric deep learning, Image-guide treatment, Pattern recognition and classification
\end{IEEEkeywords}

\section{Introduction}
\label{sec:introduction}
\IEEEPARstart{D}{espite}  the recent surge in popularity and capability of general robotic systems, automation of surgical tasks still lags behind, with no fully autonomous surgical robots on the market to date.
At the same time, surgery is a time and staff intensive process, leading to a high cognitive workload on the personnel.
While robotic systems to date act primarily as remote controlled assistive systems, improving surgeon ergonomics, enhancing dexterity and reducing physiological tremor \cite{haidegger2019motivation}, further relief can be achieved through the integration of artificial intelligence (AI) and automation into surgical workflows:
Partial automation of repetitive steps or assistance tasks holds the potential to reduce inter-procedure variability as well as reducing cognitive workload \cite{wah2025robotics_ai}.
For example, autonomous tissue exposition has been shown to reduce the cumulative activity duration required from surgical assistants \cite{younis2024holding}.
A main hindrance apart from regulatory challenges is the unpredictability and lack of discrete, segmentable entities of organic tissue in deformable and unstructured environments \cite{zhang2024sam_surgery}.
Current research circumvents this issue by relying on overly complete or externally provided knowledge of the scene for simulation or abstracting the working environment. Unfortunately, incomplete information on tissue behavior is the reality, with visual input being the only source of information routinely available during surgery. A system to seamlessly fit into the current surgical workflow in an assistive manner thus needs to rely predominantly on laparoscopic visual input.
Visual grasping approaches in realistic surgical environments however predominantly focus on the cholecystectomy procedure, reflecting a broader trend in surgical AI research \cite{carstens2025surgical_scene_understanding_review}. In this setting, the gallbladder and liver often provide visually distinguishable anatomical structures, potentially reducing the difficulty of tissue identification relative to other surgical scenarios \cite{oh2025simple_dissection}. \par
\begin{figure*}
    \centering
    \includegraphics[width=\linewidth]{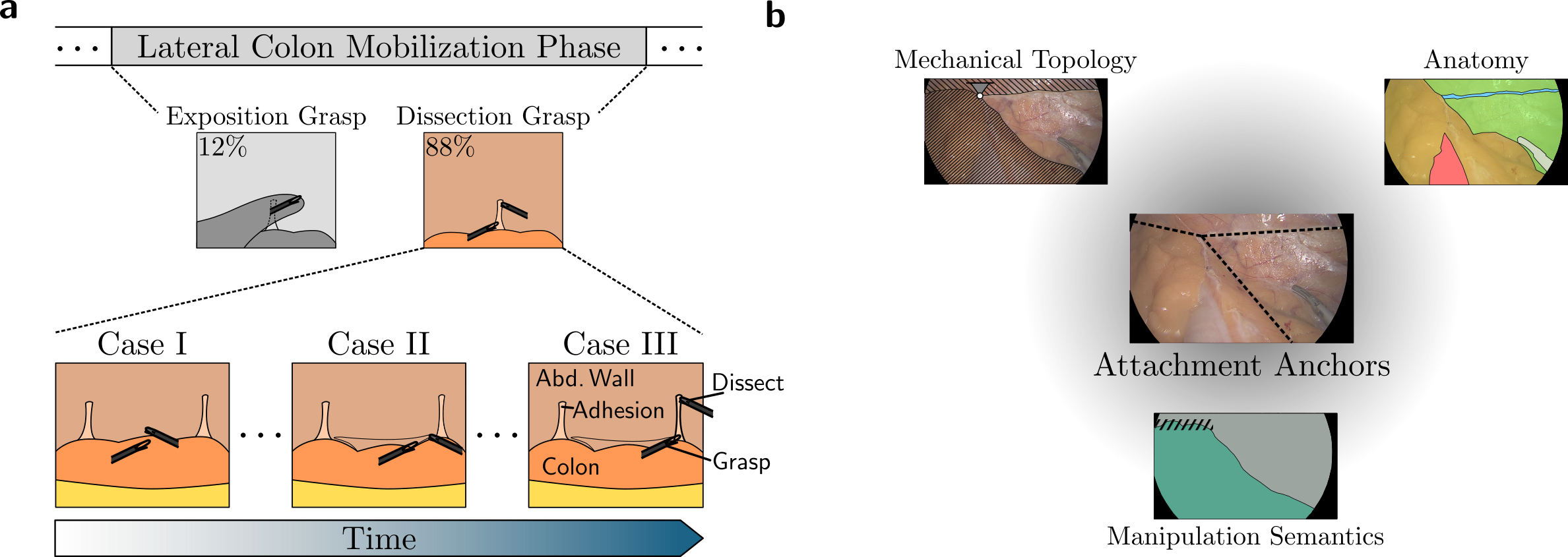}
    \caption{\textbf{\textsf{a}} The cases of the proposed attachment anchor framework originating from their surgical occurrence. The framework represents grasping maneuvers conducted for dissecting already visible tissue. These make up around 88\% of the grasping maneuvers in the lateral colon mobilization phase. \space \textbf{\textsf{b}} The attachment anchor framework derived from the mechanical topology of connected tissue, the anatomical objects and the general semantics of rigid mounting and soft graspable objects.}
    \label{fig:graphical_abstract}
\end{figure*}

%
In this article, we present attachment anchors, a local geometric representation consisting of a reference point with three associated vectors.
They define a task-specific coordinate frame, predicted directly from the laparoscopic image. 
The representation captures the structure of environments encountered during grasping maneuvers in the lateral colon mobilization phase \cite{kitaguchi2020colorectal_phases}, a phase common to all resecting colorectal surgical procedures. %
The occurrence of grasping maneuvers and their representation by the proposed framework can be seen in Fig.~\ref{fig:graphical_abstract}\textbf{\textsf{a}}.
It jointly represents the inherently present mechanics, the anatomical parts and their semantic meaning for the grasping task as seen in Fig.~\ref{fig:graphical_abstract}\textbf{\textsf{b}} and condenses the complex image into a simpler unified representation.
The framework thus emerges from a combination of multiple complementary derivations converging to the same model. \par%
We evaluate the proposed approach in both a pure statistical analysis and a learning-based grasping prediction framework.
Our analysis reveals consistent grasping patterns across surgical procedures, surgeons and surgical centers over a time span of ten years where surgical equipment, camera equipment and surgical workflows changed continuously.
Furthermore, incorporation into a visual grasping prediction pipeline demonstrates that attachment anchors can be predicted directly from laparoscopic images and statistically significantly improve grasp localization performance.
In summary, the contributions of this work are as follows:
\begin{itemize}
    \item We propose attachment anchors, a novel visual representation for the colorectal colon mobilization phase.
    \item We present a grasping point prediction framework based on attachment anchors and radial regression. We statistically validate its applicability on real-world colorectal surgical cases.
    \item We demonstrate that the proposed representation outperforms image-only baselines in visual tissue manipulation prediction and generalizes well to several changing environmental factors. 
\end{itemize}
\section{Related Work}\label{sec:related_work}
\subsection{Scene Graphs}
A major paradigm of scene simplifications for use in downstream tasks is given by scene graphs.
The core idea is the introduction of an intermediate layer of recognized entities like surgical instruments and organs, represented by either bounding boxes or segmentations.
Between these objects, relations, for example of surgical actions defined as \texttt{<subject, predicate, object>} are incorporated to create a simplified scene for downstream tasks to work on. \par%
While scene graphs provide an interpretable and rich representation, they require extensive annotation effort \cite{henriques2026scene_graphs, kolbinger2024colorectal_segmentation} and have shown to struggle with robustness under domain shift across surgeons, patients and equipment \cite{henriques2026scene_graphs}.
Moreover, existing work is largely limited to the laparoscopic cholecystectomy, limiting their applicability to anatomies with less distinct visual structure \cite{henriques2026scene_graphs}.
Annotation effort can be reduced using coarser representations such as bounding boxes \cite{murali2024latent_graph}, at the cost of downstream performance. \par%
In contrast, our work focuses on lightweight, local scene representations that avoid explicit object detection and are thus feasible to annotate at scale. 
Our work also explicitly evaluates robustness to domain-shift across surgeons, centers and equipment having continuously changed over ten years of dataset collection.
\subsection{Tissue Grasping in Simulation}
Due to the complexity of deformable environments, several works have focused on in-simulation learning of grasping and manipulation policies.
A major part of soft body physics is given by its geometric constraints. Many simulated frameworks model tissue through simplified constraints, such as fixed attachment regions or boundary conditions. This concept is generally called \emph{attachment points} and simplifies the situation.
In its simplest case, a work by Meli et al. \cite{meli2021biomedically_informed} learns a retraction trajectory given complete knowledge on the entire scene and all attachment points.
Further works add increasing complexity to such simulated environments, however predominantly focus on trajectory optimization given a predefined suitable grasping point \cite{uday2024jiggle, liang2024autopeeladhesionawaresafepeeling}. \par%
Clinical translation however proves difficult for purely simulated approaches as such detailed knowledge on tissue composition and connectivity is unrealistic in soft bodies.
Other works incorporate intra-operative visual information to solve the grasping problem in a simulated phantom.
Such works often either use registration of pre-operative data during the surgery for mapping of the organ's internals \cite{plantefeve2016preop_inop_registration}, others generate a complete 3d model from a perceived point cloud by assuming unknown properties \cite{kunz2024statistical_fitting}.
The approach by Plantef{\`e}ve et al. \cite{plantefeve2016preop_inop_registration} registers a scanned model of a liver from pre-operative MRI to the intra-operative deformable soft body using non-rigid registration.
This work however assumes an organ of distinct shape and does not extend to highly deformable and moving organs like the colon, where similar landmarks for registration are missing.
The work by Kunz et al. \cite{kunz2024statistical_fitting} directly aims to predict grasping points by assuming the gallbladder to be mirror-symmetric and maps grasping points to a fitted gallbladder-prototype.
Several of these approaches introduce significant overhead by soft-body physics solvers, ranging from 60 seconds \cite{kunz2024statistical_fitting} to 5 minutes \cite{long2025SurRoL}. Such interruption times become intractable when the camera is not centered on a single point of intervention but moves along the dissected organ gradually, as is the case in colorectal procedures as present in this article. \par%
Furthermore, our work does not impose any restrictions on the availability of pre-operative scans and operates solely on intra-operative visual input.
\subsection{Vision-based Grasping Approaches}
A system being easily integrable into current surgical workflows has obvious advantages with respect to technical adoption and extensibility.
With the laparoscopic camera being inherently in use, purely vision-based approaches severely lower this barrier and are of major research interest. \par%
A line of work focuses on extracting local visual or geometric cues relevant for tissue manipulation.
For example, the fibrous texture of dissection planes can be segmented to guide manipulation \cite{kumazu2021EUREKA, kolbinger2024colorectal_segmentation}, while stereo vision enables estimating the orientation of manipulated tissue for trajectory planning\cite{nagy2018optical_napkin_angle}.
Related approaches exploit geometric consistency by mapping observed deformations to an undeformed reference for inferring attachment locations \cite{tagliabue2021attachments}, typically relying on simplifying assumptions about tissue structure, for example in a tabletop setting. \par%
Other methods formulate grasping based on local representations of the surgical site. Liang et al. \cite{liang2025medic} model the dissection area using task-relevant directions and angles to derive grasping and retraction strategies, demonstrating that local geometric structure can be sufficient for manipulation in simplified settings. \par%
In contrast, higher-level approaches rely on semantic segmentation to guide grasping decisions.
For example, Oh et al. \cite{oh2025simple_dissection} define grasping points based on the segmented gallbladder and liver, leveraging clearly distinguishable anatomical structures.
While effective in such settings, these approaches depend on the presence of well-defined object boundaries and may be less applicable to anatomies with less distinct visual structure. \par%
Our work builds on these insights by representing the task-relevant spatial relationships of the scenery, which generalizes well as a framework as it inherently is present in the anatomy and extends to similar procedures.
\section{Problem Analysis}\label{sec:problem_analysis}
\noindent In Section~\ref{subsec:task_formulation}, we first formalize the task of surgical grasping point prediction. Section~\ref{subsec:case_discrimination} then introduces attachment anchors as a novel surgical scene representation using three representative grasping manipulation cases.
\subsection{Task Formulation}\label{subsec:task_formulation}
\noindent Let a surgical site be given by a set of interconnected tissue and organs, which the surgeon aims to separate in the course of a surgical phase.
Separation of organs is achieved by dissection, which means the repetitive and sequential cutting of connective tissue. 
For one single step this can be formulated as the dissection of a point $D$ located on the surgical site.
To achieve this task, the surgeon exposes this region by grasping and retracting the respective organ from a suitable point $G$. 
Thus, tissue exposition can be described as selecting a grasping point $G$ and direction that enable sufficient visualization of the queried dissection point $D$. 
In clinical practice, this selection is based solely on the tissue's visual appearance and the surgeon’s implicit understanding of the underlying mechanical constraints and tissue connectivity~\cite{trute2024visual_cues}.\par%
With this understanding surgeons classify the surgical site into different anatomical segments. Our approach aims to reproduce this abstracted scene understanding by formulating the task of grasping point prediction. Given an intraoperative image $I$ of the surgical site and a specified dissection point $D$, the task is to predict an appropriate grasping point $G$ on the visible tissue. Formally, we seek to learn a mapping
\begin{equation}
f_{\Phi}:(I, D) \mapsto G, 
\label{eq:model_schema}
\end{equation}
where $f_{\Phi}$ denotes a parameterized model that infers grasping points from visual input alone. We evaluate our proposed model by comparing the predicted grasping points to those selected by expert surgeons in identical visual situations. The degree to which the model reproduces the surgeons' chosen grasp locations indicates how well it captures the underlying tissue topology and the visual-mechanical reasoning required for effective tissue manipulation.
\subsection{Case Discrimination}\label{subsec:case_discrimination}
\begin{figure*}
    \centering
    \includegraphics[width=\linewidth]{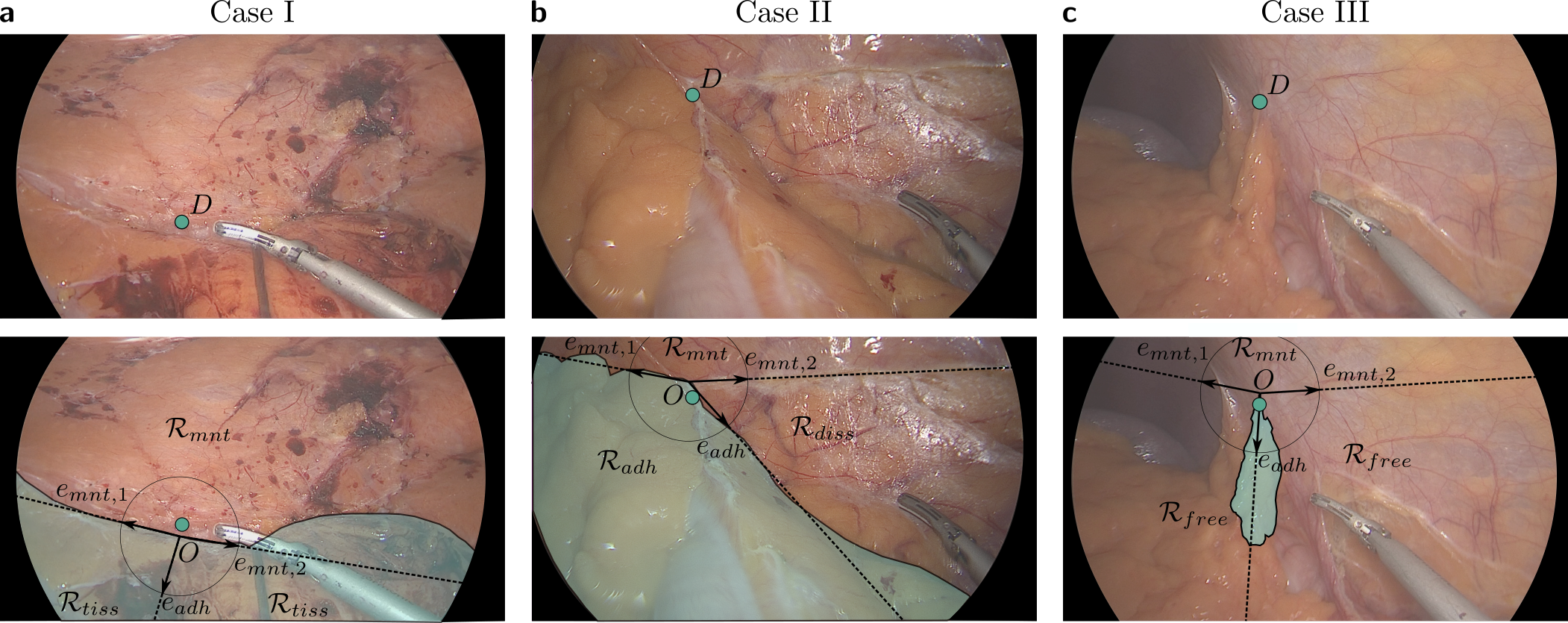}
    \caption{
    Example surgical input images illustrating the three attachment anchor cases. Each image shows the dissection point $D$ (top) and the corresponding attachment anchor representation (bottom), with the soft tissue of interest highlighted in blue. \space \textbf{\textsf{a}} Case I: The colon is fully attached to the rigid abdominal wall before colon mobilization or after change of the target anatomical region. \space \textbf{\textsf{b}} Case II: The colon is partly detached from the rigid abdominal wall during colon mobilization. The surgeon continues to dissect along the rigid tissue, progressively mobilizing the colon. \space \textbf{\textsf{c}} Case III: An adhesion strand is connecting the colon to the rigid abdominal wall, hindering further mobilization.
    }
    \label{fig:attachment_anchors}
\end{figure*}

\noindent For accurate grasping point prediction, we introduce the attachment anchor framework (see Fig.~\ref{fig:attachment_anchors}), which provides an abstract representation of the local mechanical configuration relevant to a grasping maneuver. Formally, an attachment anchor is defined in a two-dimensional polar coordinate system around its mechanical origin $O$. Originating from there, three direction vectors are defined on the unit circle $\mathbb{S}^{1}$, allowing angular relationships to be modeled independently of scale.
Two mounting vectors $\mathbf{e}_{\mathrm{mnt},1}$ and $\mathbf{e}_{\mathrm{mnt},2}$ represent attachments of the tissue to rigid or semi-rigid anatomical structures, and one adhesion vector $\mathbf{e}_{\mathrm{adh}}$ encodes the dominant direction of the soft tissue. When vectors share the same semantic role, they are indexed in clockwise order.\par%
We consider a surgical scene during the lateral colon mobilization phase, where the primary tissue components include the colon, surrounding connective tissue, and the abdominal wall. As the objective of colon mobilization is to separate the colon and its mesentery from its rigid anatomical attachments, we abstract the scene into two interacting elements, the deformable tissue to be mobilized and its rigid mounting. Based on this abstraction, we can represent the local environment relevant for manipulation by distinguishing three general retraction cases (see Fig.~\ref{fig:attachment_anchors}).\par%
\noindent \textbf{Case I: Plane adhesion.} A plane adhesion is fully attached to the rigid mounting over a wide contact area (Fig.~\ref{fig:attachment_anchors}\textbf{\textsf{a}}). The surgical objective in this configuration is to pull the adhesion away from the mounting to generate tension along the entire boundary between soft tissue and rigid structure, thereby facilitating continuous dissection along this interface. 
The plane adhesion is represented by the local boundary separating deformable tissue from the rigid mounting. As this boundary alone does not define a unique anchor location, the attachment anchor origin $O$ is defined at the boundary point closest to the dissection point $D$, ensuring an unambiguous reference.
Two vectors $\mathbf{e}_{\mathrm{mnt},1}$ and $\mathbf{e}_{\mathrm{mnt},2}$ are defined tangentially along the boundary to encode the rigid mounting, while the adhesion vector $\mathbf{e}_{\mathrm{adh}}$ points toward the deformable tissue, distinguishing it from the mounting. This formulation yields a simplified yet expressive representation of plane adhesions, capturing the distinct mechanical, semantic, and visual characteristics associated with each vector and enclosed region.\par%
The wide contact area of the adhesion results in grasping forces being distributed relatively evenly along the attachment. Consequently, feasible grasping points $G$ are less localized, as similar retraction effects can be achieved from a broad range of locations. \par%
\noindent \textbf{Case II: Adhesion wedge.} After partial dissection, a wide adhesion remains attached to the rigid mounting on one side while being detached on the other (see Fig.~\ref{fig:attachment_anchors}\textbf{\textsf{b}}). The dissection point $D$ is located along the remaining attachment between the adhesion and the rigid structure, indicating that dissection should continue at this interface. 
The surgical objective is to grasp the mobilized tissue and open up the wedge towards the rigid body in a hinge-like manner. %
We thus model the adhesion wedge as a mechanical hinge between the attached soft tissue and the tissue that has already been dissected. The hinge location is defined at the intersection of both attachment states. A unit vector $\mathbf{e}_{\mathrm{adh}}$ follows the detached edge of the adhesion tangentially and represents the direction in which the hinge opens. Two mounting vectors, $\mathbf{e}_{\mathrm{mnt},1}$ and $\mathbf{e}_{\mathrm{mnt},2}$, encode the current and former attachment to the rigid mounting, respectively. This gives rise to the triangular shape of the deformable tissue. Beyond their mechanical interpretation, the vectors enclose regions with distinct semantic and visual characteristics. Region $\mathcal{R}_{\mathrm{adh}}$ between $\mathbf{e}_{\mathrm{adh}}$ and $\mathbf{e}_{\mathrm{mnt},1}$ corresponds to the manipulable soft tissue. Region $\mathcal{R}_{\mathrm{mnt}}$ between $\mathbf{e}_{\mathrm{mnt},1}$ and $\mathbf{e}_{\mathrm{mnt},2}$ captures the rigid mounting which tends to have different texture and coloring. Region $\mathcal{R}_{\mathrm{diss}}$ between $\mathbf{e}_{\mathrm{mnt},2}$ and $\mathbf{e}_{\mathrm{adh}}$ captures the open hinge, which is visually distinct due to its marks from previous dissection. It also represents the semantic meaning of opening up for target fulfillment.\par%
\noindent \textbf{Case III: Adhesion strand.} A single, narrow adhesion strand connects the deformable tissue to the rigid mounting, with the dissection point $D$ located on the adhesion itself (see Fig.~\ref{fig:attachment_anchors}\textbf{\textsf{c}}).
The adhesion is represented by a unit vector $\mathbf{e}_{\mathrm{adh}}$ aligned with the strand’s centerline and originating at the visually identifiable attachment point $O$ on the rigid mounting farthest from the deformable tissue. The rigid mounting is encoded by the two unit vectors $\mathbf{e}_{\mathrm{mnt},1}$ and $\mathbf{e}_{\mathrm{mnt},2}$, capturing its orientation and perspective. While the adhesion vector specifies the location and orientation of the adhesion, the enclosed region $\mathcal{R}_{\mathrm{mnt}}$ spanned by the mounting vectors represents the visible features of the rigid mounting region.
To generate linear tension along the strand, the grasping point $G$ is selected along its linear extension.
\section{Methodology}\label{sec:methodology}
\begin{figure}[b]
    \centering
    \includegraphics[width=\linewidth]{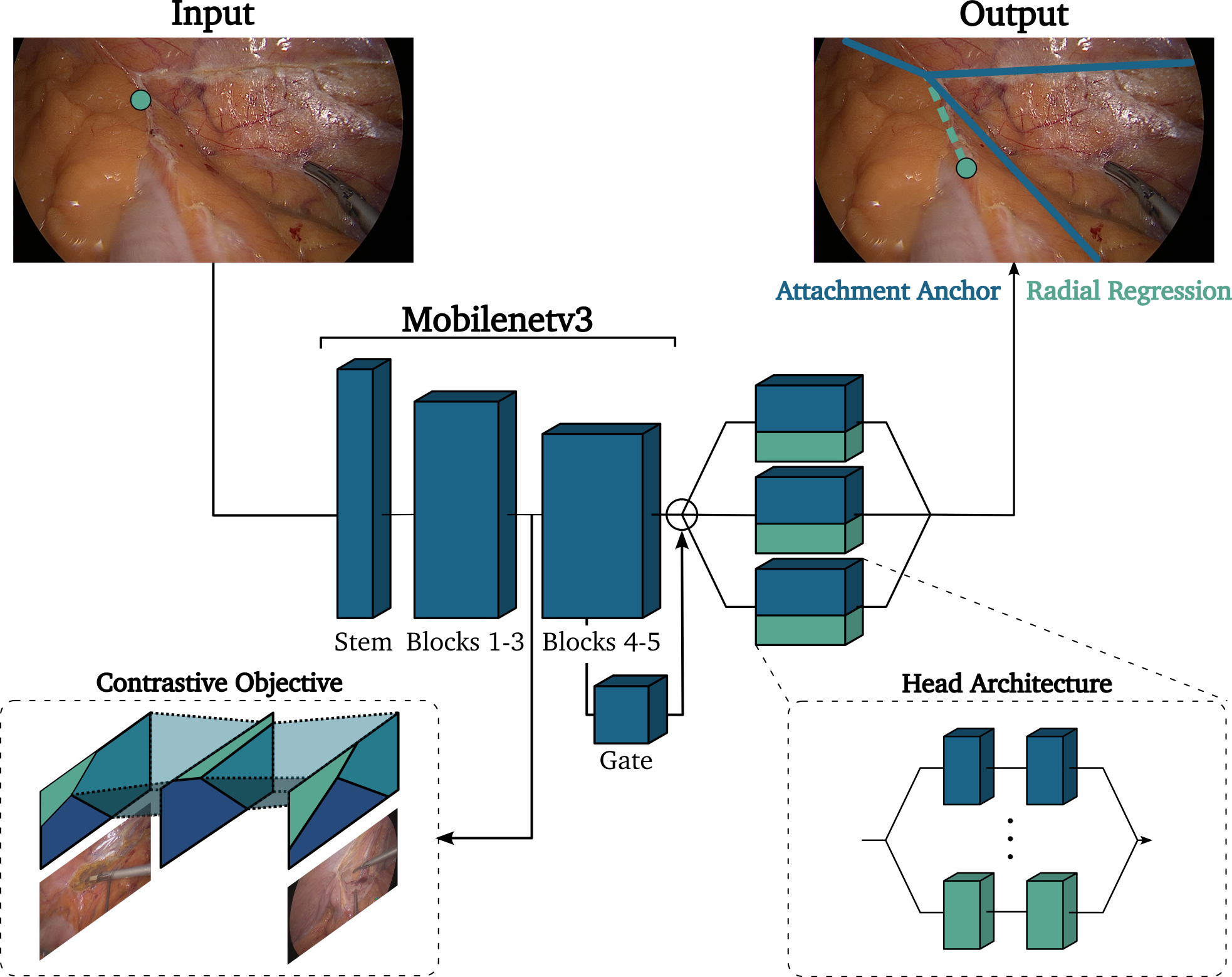}
    \caption{The architecture of the proposed model. The attachment anchor encoder $\phi_A$ (blue) extracts the anchor representation, which is then used by the grasping point decoder $\phi_G$ (cyan).}
    \label{fig:pipeline}
\end{figure}
\noindent In Section~\ref{subsec:encoder}, we first describe the attachment anchor encoder, which extracts a structured anchor representation from an input image and query dissection point. In Section~\ref{subsec:decoder}, we then present the task-specific grasping point decoder that predicts grasp locations based on the learned attachment topology. Finally in Section~\ref{subsec:dataset}, we describe the dataset used for training and evaluation.
\subsection{Attachment Anchor Encoder}\label{subsec:encoder}
\noindent The attachment anchor representation $A$ of an image $I$ and dissection target $D$ is obtained by passing it through our attachment anchor encoder $\Phi_A$:
\begin{equation}
A = \Phi_A(I, D).
\label{eq:aa_formula}
\end{equation}
\noindent The encoder $\Phi_A$ is implemented as a YOLOv8-based \cite{redmon2016yolo} object detection architecture with a \textit{MobileNetV3-small} \cite{howard2019mobilenetv3} backbone. To balance spatial resolution and semantic richness, we extract intermediate feature maps from layer $\textit{blocks.3}$, which serve as the basis for medium-resolution feature learning.
A custom detection head is appended for attachment anchor prediction. This head performs keypoint-based detection, identifying both the mechanical origin $O$ of an attachment anchor and its three defining directional vectors $\mathbf{e}_{\mathrm{adh}}$, $\mathbf{e}_{\mathrm{mnt},1}$, and $\mathbf{e}_{\mathrm{mnt},2}$. To account for the different anchor configurations introduced in Section~\ref{subsec:case_discrimination}, we employ one prediction head per anchor type. A discrete gate $o_{\mathrm{gate}} \in \{0, 1, 2\}$ classifies the responsible head for each activated grid cell. 
Each prediction head outputs an origin-offset $c \in [-1,1]^2$, localizing the respective attachment anchor within the activated grid cell, along with three two-dimensional anchor-vectors $\mathbf{e}_i \in [0,1]^2, i \in \{0,1,2\}$, encoding the orientation of the attachment anchor.\par%
To ensure feature-consistency across semantically similar regions, we further leverage the intermediate feature maps for contrastive representation learning at medium resolution. Specifically, we apply an InfoNCE-Loss \cite{oord2018infonce} to enforce similarity between feature embeddings corresponding to semantically equivalent regions. We consider regions $\mathcal{R}_{\mathrm{mnt}}$ as semantically equivalent throughout all cases.
Likewise, multiple occurring regions $\mathcal{R}_{\mathrm{free}}$ (and $\mathcal{R}_{\mathrm{tiss}}$) are treated as semantically equivalent, reflecting symmetry. 
The attachment encoder pipeline is shown in Fig.~\ref{fig:pipeline}. 
\subsection{Grasping Point Decoder}\label{subsec:decoder}
\noindent The learned attachment anchor representation $A$ is subsequently leveraged to predict the grasping point. To this end, we introduce a grasping point detection decoder $\Phi_G$, which extends on the pretrained attachment prediction head with situation-aware grasping point regression. Given an attachment anchor representation $A$ encoded by $\Phi_A$, the decoder $\Phi_G$ predicts the grasping point in a local, anchor-centric coordinate system. Specifically, the grasping location is parameterized in polar coordinates relative to the predicted anchor origin. The decoder regresses an offset direction
\begin{equation}
\phi_{\mathrm{rel}} \in \mathbb{R}^2, \; \; \; \; \; \; \; \lVert \phi_{\mathrm{rel}} \rVert_2 = 1,
\label{eq:encoder}
\end{equation}
which represents a direction vector on the unit circle and is defined relative to the adhesion direction vector $\mathbf{e}_{\mathrm{adh}}$. This formulation enforces rotational consistency between the anchor orientation and the predicted grasp direction. In addition, the decoder predicts a relative radial distance $r_{\mathrm{rel}} \in [0,1]$ obtained via a sigmoid activation and scaled by the diagonal length of the input image, thereby yielding the final grasping point location
\begin{equation}
G = O_{\mathrm{anchor}} + r_{\mathrm{rel}} \cdot \mathrm{diag}(I) \cdot \phi_{\mathrm{rel}}
\label{eq:decoder}
\end{equation}
By conditioning grasp prediction on anchor position, orientation, and anchor type, the decoder enables situation-aware grasping, with each anchor-specific prediction head learning a distinct solution manifold tailored to its corresponding retraction case. The resulting architecture, consisting of the anchor-pretrained encoder $\Phi_A$ and the task-specific grasp decoder $\Phi_G$ with radial regression, is referred to as RadAttAnc.
\subsection{Dataset}\label{subsec:dataset}
\begin{figure}[tb]
\centering
\includegraphics[width=0.35\textwidth]{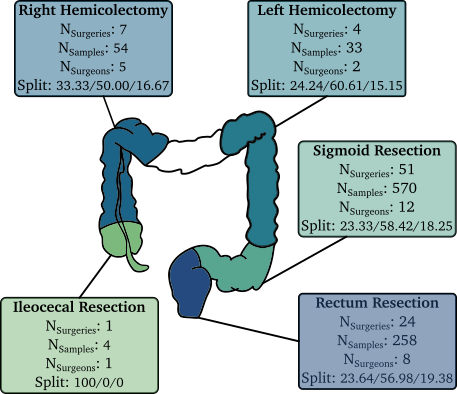}
\caption{Dataset composition showing colorectal surgical procedures stratified by anatomical region along the colon.}
\label{fig:dataset}
\end{figure}

\noindent For learning the encoder $\Phi_A$ and decoder $\Phi_G$, we use a self-curated in-house dataset of laparoscopic colorectal surgeries. The dataset comprises 90 colorectal surgical interventions collected at the TUM University Hospital between 2015 and 2025. Data collection was conducted under ethical approval from the Ethics Committee of the TUM University Hospital (approval number: 337/21 S), and written informed consent was obtained from all patients. 
The dataset consists of laparoscopic colorectal procedures spanning five anatomical subregions along the colon, as illustrated in Fig.~\ref{fig:dataset}. Specifically, ileocecal resections involve mobilization and removal of the terminal ileum and cecum. Right hemicolectomies extend this resection to the ascending colon and hepatic flexure, while left hemicolectomies address the descending colon and splenic flexure. Sigmoid resections focus on the mobilization and removal of the sigmoid colon, and rectal resections entail deep pelvic dissection of the rectum. In addition, the dataset contains three proctocolectomies, in which the entire colon is resected.\par%
The surgeries were performed by 15 different surgeons with diverse clinical backgrounds, resulting in a dataset reflecting a wide range of operative styles and anatomical variability. Colon mobilization is a critical phase common to all resecting colorectal procedures. Therefore, the dataset is built around grasping maneuvers occurring during the lateral colon mobilization phase. 
The dataset consists of image samples of individual grasping maneuvers, each extracted at a time point immediately preceding tissue grasping where the surgeon has an unobstructed view of the intended dissection target.
This is the case in around 88\% of occurring grasps, with obstructing anatomies occurring in around 12\% of cases and being preventable by adequate patient positioning.
\subsection{Evaluation Metrics}\label{subsec:evaluation_metrics}
\noindent Model performance is evaluated in terms of grasping point localization using \textit{Precision@6\%}. A prediction is considered correct if it lies within a circular region centered on the ground-truth grasping point. Given the $7\times 7$ output grid of the YOLO backbone, the radius of this region is set to one-seventh of the image size. This corresponds to a positive area covering approximately $6.4\%$ of the input image. Since attachment anchors tend to occur close to the query dissection point, non-maximum suppression yields a single attachment anchor and thus a single grasp prediction for most samples. Thus Recall and F1-score are omitted because they do not differ significantly across models.\par 
As a baseline, we use a standard YOLOv8 architecture adapted for keypoint detection. This visual baseline architecture directly predicts grasping points from images without any anchor-based representation, following the setup described in Section~\ref{subsec:encoder}. A second variant incorporates attachment anchors as an auxiliary pretraining objective while retaining the same absolute grasp localization formulation (AbsAttAnc). Finally, we evaluate an architecture explicitly leveraging the attachment anchor representation and employing radial regression for grasp localization relative to its predicted attachment anchor (RadAttAnc).
\section{Experiments and Results}\label{sec:results}
\noindent In this section, we evaluate the impact of attachment anchors on surgical grasping point prediction. In Section~\ref{subsec:exp_statistical}, we first demonstrate the statistical simplification on the grasping problem enabled by the attachment anchor framework and analyze the predictability of the attachment anchors from surgical images (Section~\ref{subsec:exp_anchor_prediction}).
We then assess their effect on the grasping point prediction task using attachment anchor knowledge (Section~\ref{subsec:exp_attachment_anchors}), followed by generalization studies on unseen surgical procedures (Section~\ref{subsec:exp_unseen_surgeries}), unseen surgeons (Section~\ref{subsec:exp_unseen_surgeons}) and an unseen clinical center (Section~\ref{subsec:exp_unseen_centers}). 
\subsection{Statistical Analysis}\label{subsec:exp_statistical}
\begin{figure*}
    \centering
    \includegraphics[width=\textwidth]{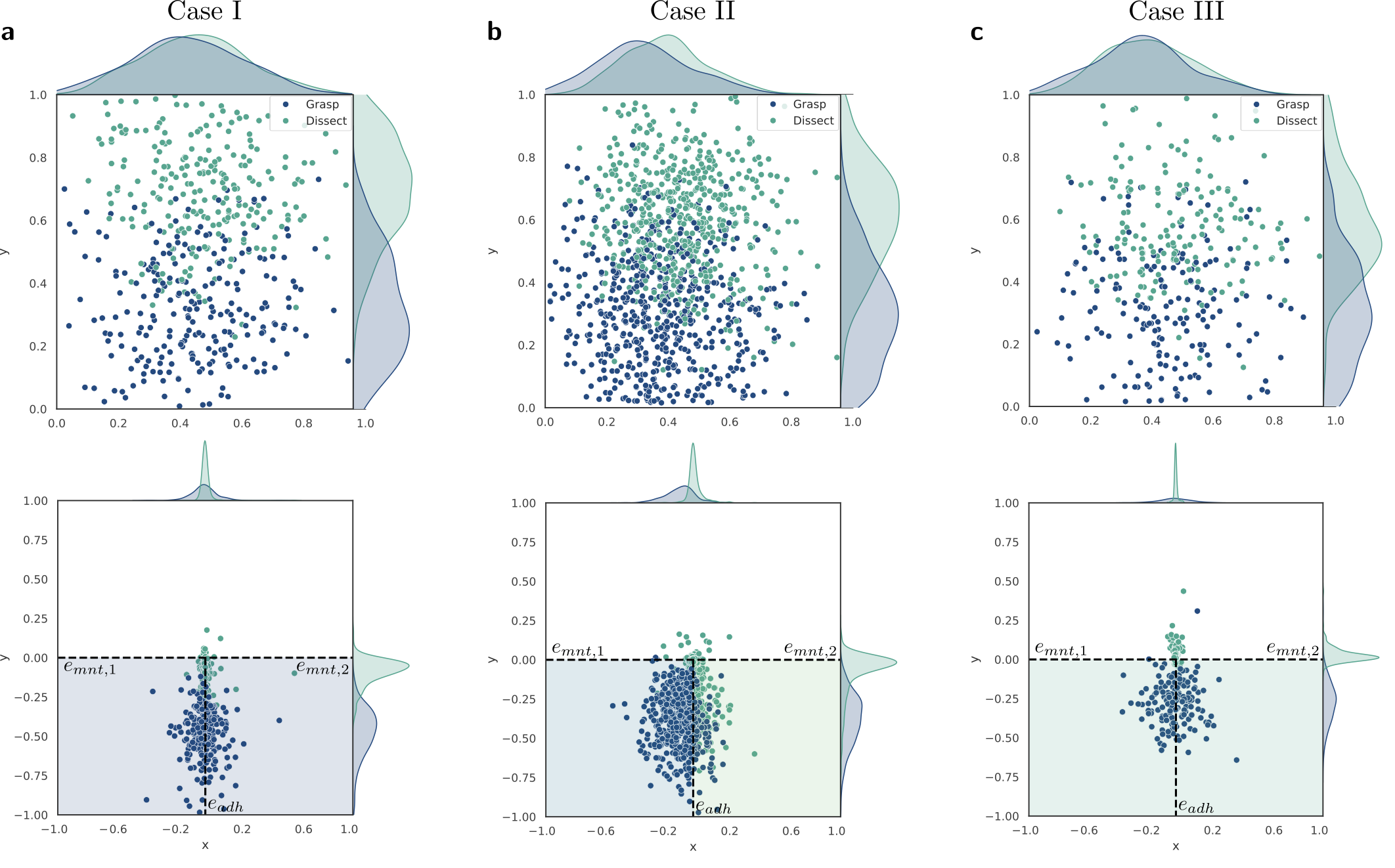}
    \caption{The statistical distribution of grasping and dissection points per case (top) and after normalization with their respective attachment anchor origin and orientation (bottom).}
    \label{fig:gd_dist}
\end{figure*}

\noindent We analyze the effect of attachment anchors on the grasping task by examining the spatial distributions of grasping and dissection points normalized to the attachment anchor representation (Fig.~\ref{fig:gd_dist}). All samples are grouped by attachment anchor case and superimposed after a normalization procedure aligns the anchors to a common reference frame. Specifically, each anchor is translated to a shared origin while its defining vectors are warped to fixed orientations. Vectors
$\mathbf{e}_{\mathrm{mnt},1}$ and $\mathbf{e}_{\mathrm{mnt},2}$ are aligned horizontally. Vector $\mathbf{e}_{\mathrm{adh}}$ is normalized vertically downward. This warping transformation keeps ratios between the anchor directions and radial distance to the attachment anchor's origin intact.\par%
A closer inspection of the normalized distributions in Fig.~\ref{fig:gd_dist} reveals case-specific patterns.
In case~I, dissection points are tightly clustered below the anchor origin along the adhesion direction $\mathbf{e}_{\mathrm{adh}}$, reflecting the mechanical origin of the adhesion. Correspondingly, grasping points form a linear extension along the same direction. Case~II, which is inherently asymmetric, is normalized such that region $\mathcal{R}_{\mathrm{adh}}$ lies in the lower-left quadrant. Dissection points concentrate along $\mathbf{e}_{\mathrm{adh}}$ and $\mathbf{e}_{\mathrm{mnt},1}$, consistent with continued dissection along the remaining adhesion, while grasping points predominantly lie within $\mathcal{R}_{\mathrm{adh}}$ near $\mathbf{e}_{\mathrm{adh}}$ to maintain exposure of the dissection target. In case~III, dissection points tend to cluster near the boundary between soft tissue and mounting, whereas grasping points exhibit substantially higher variability. This reflects the evenly distributed attachment of plane adhesions, where multiple grasping points yield similar tissue responses. This leads surgeons to prioritize ergonomic considerations. %
Table~\ref{tab:std_data} reports the standard deviations (SDs) of the resulting distributions of the absolute grasping points \textit{(Absolute)}, grasping points expressed relative to their dissection targets \textit{(Relative)}, and grasping points after attachment anchor-based normalization. Across all cases, anchor normalization yields a statistically significant reduction in standard deviation along the x-axis in a one-sided pairwise t-test between relative and attachment anchor-normalized representation. 
\begin{table}[ht]
\caption{Standard deviations of grasping and dissection point distributions across absolute coordinates, relative-to-dissection coordinates, and attachment anchor-based normalized representations.}
\centering
\begin{tabular}[t]{lcccc}
\toprule
& \multicolumn{2}{c}{Grasp} & \multicolumn{2}{c}{Dissect} \\
\textbf{Type} & \textbf{std(x)} (\%) & \textbf{std(y)} (\%) & \textbf{std(x)} (\%) & \textbf{std(y)} (\%) \\
\midrule
\textit{Absolute}       &               17.81 &              17.09 & 16.45 & 17.22 \\
\midrule
Case I    & 18.29 & 16.91 & 17.54 & 16.19 \\
Case II    & 16.87 & 17.11 & 15.32 & 17.01 \\
Case III    & 18.05 & 17.41 & 17.27 & 16.24 \\
\midrule
\textit{Relative}       &               13.53 &              15.74 & - & - \\
\midrule
Case I    & 11.77 & 14.03 & - & - \\
Case II    & 13.19 & 15.94 & - & - \\
Case III    & 15.52 & 14.51 & - & - \\
\midrule
\multicolumn{2}{l}{\textit{Attachment Anchors}} &&& \\
\midrule
Case I         &       $\textbf{8.57}\textrm{*}$ &               15.69 & 1.40 & 8.03 \\
Case II         &      $\textbf{10.08}^\dagger$ &              16.44 & 4.63 & 14.27 \\
Case III         &      $\textbf{11.66}^\dagger$ &              12.85 & 1.16 & 4.24 \\
\bottomrule
\tiny{(one-sided pairwise t-test)}\\
\multicolumn{3}{l}{\tiny{$^\dagger\;\textrm{p}<0.01$}\quad\quad\quad\quad\tiny{$\textrm{*}\;\textrm{p}<0.05$}}\\
\end{tabular}
\label{tab:std_data}
\end{table}
\subsection{Attachment Anchor Prediction}\label{subsec:exp_anchor_prediction}
\noindent We next assess the quality of predicting an attachment anchor suitably representing the grasping environment queried by a dissection point $D$ on an input image $I$. 
Quantitative results are reported in Table~\ref{tab:anchor_prediction}, obtained using 5-fold cross-validation.
The folds are separated on a surgery-level to balance the fold-wise distribution of attachment anchor cases and the total sample count. \par%
Incorporating regional consistency as defined by attachment anchors through contrastive feature learning leads to a substantial improvement in attachment anchor localization performance. Specifically, Precision@6\% increases by $9.60\%$ accompanied by a marked reduction in localization error, indicating more accurate anchor origin prediction.
The prediction accuracy of the anchor type improves only marginally, suggesting that the primary benefit of the contrastive objective lies in learning a more discriminative spatial representation rather than enhanced categorical discrimination. 
\begin{table*}[ht]
\caption{Attachment anchor prediction performance (mean $\pm$ sd $\%$) under 5-fold cross-validation. Results compare a purely image-based keypoint detector (Baseline) with the same architecture augmented by contrastive learning.}
\centering
\begin{tabular}[t]{lcccccc}
\toprule
& \textbf{Precision@6\%} $\uparrow$ & \textbf{RMSE} $\downarrow$ & \textbf{Type Precision} $\uparrow$ & \multicolumn{3}{c}{\textbf{RMSE} (rad) $\downarrow$} \\
 
\textbf{Model} & \multicolumn{2}{c}{Origin $O$} & Type $o_{\mathrm{gate}}$ & $\mathbf{e}_{\mathrm{adh}}$ & $\mathbf{e}_{\mathrm{mnt},1}$ & $\mathbf{e}_{\mathrm{mnt},2}$ \\
\midrule
Baseline      & $78.24  \pm 2.62$ & $12.91 \pm 4.81$ & $64.50 \pm 5.64$ & $0.55 \pm 0.04$ & $0.63 \pm 0.04$ & $0.65 \pm 0.09$ \\
\midrule
+\,Contrastive & $\textbf{87.84} \pm \textbf{3.40}^\dagger$ & $\textbf{9.75}  \pm \textbf{1.66}$ & $\textbf{67.17} \pm \textbf{7.84}$ & $\textbf{0.48} \pm \textbf{0.03}\textrm{*}$ & $\textbf{0.57} \pm \textbf{0.04}\textrm{*}$ & $\textbf{0.60} \pm \textbf{0.04}$ \\
\bottomrule
\multicolumn{3}{l}{\tiny{$^\dagger\;\textrm{p}<0.01$}\quad\quad\quad\quad\tiny{$\textrm{*}\;\textrm{p}<0.05$}}\\
\end{tabular}
\label{tab:anchor_prediction}
\end{table*}
\subsection{Experiments Using Attachment Anchors}\label{subsec:exp_attachment_anchors}
\noindent We next evaluate the impact of attachment anchor representations on the grasping point prediction task. Prediction performance is compared across the proposed model variants using 5-fold cross-validation with evenly distributed splits, following the same protocol as in Section~\ref{subsec:exp_anchor_prediction}. Quantitative results are summarized in Table~\ref{tab:grasp_results}.\par%
Incorporating attachment anchor information yields a substantial and statistically significant improvement in grasp localization performance. Compared to the vision-only baseline (Baseline), the inclusion of attachment anchor pretraining (AbsAttAnc) increases Precision@6\% by $12.10\%$, while also reducing localization error as measured by RMSE. Adding an explicit radial grasping representation to the model (RadAttAnc) results in a slight increase of $1.30\%$ in Precision@6\%. However, this improvement is not statistically significant. This suggests that the primary performance gain stems from the attachment anchor representation itself rather than from the specific grasp parameterization.
\begin{table}[tb]
\caption{Grasping point prediction performance (mean $\pm$ sd $\%$) under 5-fold cross-validation. Results compare vision-only prediction with models incorporating attachment anchor representations.}
\centering
\begin{tabular}[t]{lcc}
\toprule
\textbf{Model} & \textbf{Precision@6\%} $\uparrow$ & \textbf{RMSE} $\downarrow$ \\
\midrule
Baseline      & $37.80  \pm 2.44$ & $18.63 \pm 0.84$ \\
\midrule
AbsAttAnc     & $49.90 \pm 4.60$ & $\textbf{14.77}  \pm \textbf{1.12}^\dagger$ \\
RadAttAnc     & $\textbf{51.20} \pm \textbf{3.70}^\dagger$ & $15.00 \pm 1.44$ \\
\bottomrule
\multicolumn{3}{l}{\tiny{$^\dagger\;\textrm{p}<0.01$}\quad\quad\quad\quad\tiny{$\textrm{*}\;\textrm{p}<0.05$}}\\
\end{tabular}
\label{tab:grasp_results}
\end{table}
\subsection{Generalization to Unseen Surgical Procedures} \label{subsec:exp_unseen_surgeries}
\noindent We evaluate the ability of the proposed approach to generalize to previously unseen surgical procedures. 
In this setting, both attachment anchor prediction and subsequent grasping point prediction are evaluated exclusively on surgical procedures not observed during training. This evaluation is particularly challenging, as the dataset spans multiple anatomical subregions of the colon (see Fig.~\ref{fig:dataset}), each exhibiting distinct visual appearances and tissue configurations.\par%
Quantitative results are reported in Table~\ref{tab:unseen_surgeries}. Incorporating attachment anchor representations substantially improves performance across all unseen surgical categories. Notably, the largest performance gain of $16.44\%$ is observed for right hemicolectomy procedures, where the vision-only baseline struggles due to pronounced visual and anatomical differences compared to the training data. In contrast, left hemicolectomy procedures exhibit a smaller but still significant improvement of $9.26\%$, which can be attributed to their greater anatomical similarity to sigmoid resections encountered during training.\par%
Overall, attachment anchors consistently enhance generalization performance across all surgical types, with statistically significant improvements in every surgical procedure ($p \leq 0.012$, unpaired t-test). These results indicate that the proposed representation captures local scene-level structural information that is transferable across diverse colorectal surgical procedures.
\begin{table*}[!ht]
\caption{Grasping point prediction performance (mean $\pm$ sd $\%$) on unseen surgical procedures across 5 evaluation runs.}
\centering
\begin{tabular}[t]{lccccc}
\toprule
\multirow{3}{*}{\textbf{Model}} 
& \multicolumn{4}{c}{\textbf{Precision@6\% on unseen surgery type} $\uparrow$} 
& \multirow{3}{*}{\textbf{Mean} (\%)} \\
\cmidrule(lr){2-5}
& \textbf{Sigmoid}
& \textbf{Rectum}
& \textbf{Hemicol. left}
& \textbf{Hemicol. right}
&  \\
\midrule
Baseline      & $30.66 \pm 2.77$ &  $41.60 \pm 3.01$ & $27.04 \pm 4.32$ & $33.89 \pm 10.26$ & 33.30 \\
\midrule
AbsAttAnc     & $43.39 \pm 1.04$ & $52.06 \pm 1.86$ & $\textbf{36.30} \pm \textbf{1.48}^\dagger$ & $\textbf{50.33} \pm \textbf{4.90}\textrm{*}$ & 45.52 \\
RadAttAnc     & $\textbf{44.10} \pm \textbf{1.54}^\dagger$ & $\textbf{52.22} \pm \textbf{1.65}^\dagger$ & $35.40 \pm 2.44$ & $48.85 \pm 4.45$ & 45.14 \\
\bottomrule
\multicolumn{3}{l}{\tiny{$^\dagger\;\textrm{p}<0.01$}\quad\quad\quad\quad\tiny{$\textrm{*}\;\textrm{p}<0.05$}}\\
\end{tabular}
\label{tab:unseen_surgeries}
\end{table*}
\subsection{Generalization to Unseen Surgeons} \label{subsec:exp_unseen_surgeons}
\noindent The dataset comprises grasping actions recorded during routine clinical practice. This introduces inevitable surgeon-specific biases because only one grasping action is annotated as correct for each situation. To assess the extent to which the proposed model generalizes beyond individual ergonomic preferences, we evaluate performance on previously unseen operating surgeons.\par%
Due to an imbalance in the number of samples per surgeon, surgeons are grouped to ensure that each group accounts for at least 15\% of the dataset, enabling statistically meaningful evaluation. The surgeon occurring the most in this dataset is not ablated. Generalization across surgeon groups provides insight into whether the model captures task-relevant grasping principles rather than overfitting to individual grasping styles. Quantitative results are reported in Table~\ref{tab:unseen_surgeons}.\par%
Without attachment anchor representations, the vision-only baseline exhibits limited generalization to unseen surgeons, achieving a mean Precision@6\% of $37.76\%$. Incorporating attachment anchors substantially improves robustness, increasing mean precision to $50.14\%$. The radial grasping formulation yields comparable performance ($49.95\%$). 
\begin{table*}[!ht]
\caption{Grasping point prediction performance (mean $\pm$ sd $\%$) on unseen surgeons across 5 evaluation runs. Surgeons are grouped to ensure sufficient sample sizes per group.}
\centering
\begin{tabular}[t]{lccccccc}
\toprule
\multirow{3}{*}{\textbf{Model}} 
& \multicolumn{6}{c}{\textbf{Precision@6\% on unseen surgeon identifier} $\uparrow$} 
& \multirow{3}{*}{\textbf{Mean} (\%)} \\
\cmidrule(lr){2-7}
& \textbf{1}
& \textbf{2}
& \textbf{3}
& \textbf{4, 5, 6}
& \textbf{7, 8, 9}
& \textbf{10, 11, 12, 13, 14}
&  \\
\midrule
Baseline      & $43.05 \pm 2.76$ &  $34.71 \pm 2.00$ & $43.56 \pm 5.35$ & $33.25 \pm 5.22$ & $33.72 \pm 2.65$ & $38.28 \pm 6.67$ & 37.76 \\
\midrule
AbsAttAnc     & $\textbf{56.27} \pm \textbf{3.62}^\dagger$ & $38.82 \pm 2.73$ & $\textbf{53.97} \pm \textbf{1.49}^\dagger$ & $53.18 \pm 3.97$ & $\textbf{48.57} \pm \textbf{1.43}^\dagger$ & $\textbf{50.00} \pm \textbf{3.45}^\dagger$ & 50.14 \\
RadAttAnc     & $55.17 \pm 4.74$ & $\textbf{40.68} \pm \textbf{3.50}^\textrm{*}$ & $51.77 \pm 1.32$ & $\textbf{55.82} \pm \textbf{4.59}^\dagger$ & $48.20 \pm 3.35$ & $48.05 \pm 2.71$ & 49.95 \\
\bottomrule
\tiny{$^\dagger\;\textrm{p}<0.01$}\quad\quad\quad\quad
\tiny{$\textrm{*}\;\textrm{p}<0.05$}\\
\end{tabular}
\label{tab:unseen_surgeons}
\end{table*}
\subsection{Generalization to an Unseen Surgical Center} \label{subsec:exp_unseen_centers}
\noindent To further evaluate robustness to variations in imaging systems, lighting conditions and surgical instrumentation, we conduct an additional experiment in which all data from our training set is evaluated on previously unseen data from the Heidelberg Colorectal Dataset \cite{maierhein2021heidelberg_colorectal_dataset}, also introducing variations on procedural habits and trocar positioning and thus variations in camera viewpoints and anatomy preparation.\par%
For this evaluation, the sigmoid resections and rectal resections of the dataset were extracted and re-annotated with attachment anchors and occurred grasping points. Grasping points were discarded if there was no clear line-of-sight on the grasped tissue before dissection or instruments were obscuring the area of interest.
Furthermore, the Heidelberg dataset often used two graspers during the interventions, with one being operated by a surgical assistant, highly affecting grasping behavior. Thus, such grasps were also left out of consideration.
In the end, 7 sigmoid resections and 10 rectum resections were annotated with 62 and 65 grasping actions, respectively. \\
Table~\ref{tab:unseen_center} lists the evaluation scores of models trained on the initial dataset using the Heidelberg dataset as a single-center holdout set.
The results demonstrate that the proposed representation maintains strong predictive performance despite the shift in center-specific characteristics, indicating improved invariance to institutional differences in data acquisition and surgical workflow.
\begin{table*}[!ht]
\caption{Grasping point prediction performance (mean $\pm$ sd $\%$) on an unseen surgical center across 5 evaluation runs.}
\centering
\begin{tabular}[t]{lcccc}
\toprule
\multirow{3}{*}{\textbf{Model}} 
& \multicolumn{2}{c}{\textbf{Sigmoid}} 
& \multicolumn{2}{c}{\textbf{Rectum}} \\
\cmidrule(lr){2-5}
& \textbf{Precision@6\%} $\uparrow$
& \textbf{RMSE} $\downarrow$
& \textbf{Precision@6\%} $\uparrow$
& \textbf{RMSE} $\downarrow$ \\
\midrule \\
Baseline     & $56.25 \pm 3.29$ & $13.26 \pm 1.26$ & $58.22 \pm 6.53$ & $12.56 \pm 1.48$ \\
\midrule
AbsAttAnc     & $\textbf{68.07} \pm \textbf{2.92}^\dagger$ & $\textbf{10.69} \pm \textbf{1.29}\textrm{*}$ & $\textbf{68.25} \pm \textbf{9.42}$ & $\textbf{10.35} \pm \textbf{1.37}\textrm{*}$ \\
RadAttAnc     & $66.87 \pm 5.92^\dagger$ & $12.05 \pm 1.40$ & $67.32 \pm 5.20\textrm{*}$ & $11.21 \pm 1.84$ \\
\bottomrule
\tiny{$^\dagger\;\textrm{p}<0.01$}\quad\quad\quad\quad
\tiny{$\textrm{*}\;\textrm{p}<0.05$}\\
\end{tabular}
\label{tab:unseen_center}
\end{table*}
\section{Discussion}\label{sec:discussion}
\noindent Our results demonstrate that attachment anchors significantly reduce uncertainty for the task of surgical grasping point prediction. Assuming perfect attachment anchor prediction, we show that the grasping problem becomes substantially more constrained, highlighting their theoretical relevance as an intermediate representation. Furthermore, we demonstrate that attachment anchors can be reliably predicted from surgical images.\par%
A central finding is that attachment anchors enable strong generalization capabilities. These performance gains are particularly pronounced in slightly out-of-distribution settings, such as unseen surgical procedures with distinct anatomical characteristics. For instance, in left and right hemicolectomies, where the visual appearance differs greatly due to the presence of structures like the liver, attachment anchors notably enhance robustness (see Table~\ref{tab:unseen_surgeries}). These results suggest that the proposed representation effectively mitigates uncertainty arising from anatomical variability by focusing on local, mechanically grounded scene structure rather than global appearance.\par%
The generalization-improvement to unseen surgeons further supports the idea that attachment anchors capture grasping principles driven by the task rather than surgeon-specific ergonomic preferences (see Table~\ref{tab:unseen_surgeons}). Further shown on data from the Heidelberg surgical center, grasping points are reliably predicted despite changes in clinical workflows like trocar positioning and tissue preparation practices (see Table~\ref{tab:unseen_center}). Due to a smaller sample size of available data, however, the findings on the Heidelberg center are only a qualitative measure. \par%
Beyond improving performance and generalization in grasping point localization, attachment anchors provide an explicit representation of how the model perceives grasping-relevant situations and offer a verifiable representation of the scenery. This verifiability is particularly valuable in safety-critical domains such as surgery. %
Erroneous anchor predictions can be identified by human oversight and inspected or manually corrected.\par%
Although attachment anchors offer an effective representation of situations relevant to grasping, their applicability is subject to several practical limitations. The main dataset used in this study has inherent biases, such as imbalances among surgeons and procedures.
We aim to mitigate the effects through careful experimental design, statistical evaluation and evaluation on an unknown clinical center.
However, the manual operation of the laparoscopic camera inherently introduces a significant central bias in grasping and dissection point locations(see Fig.~\ref{fig:gd_dist}).
In addition, the analysis is restricted to samples in which both the grasping and dissection points are visible. This reduces the size of the available dataset and excludes situations with partial occlusions occurring in around 12\% of cases. Extending the framework to handle hidden or temporally inferred targets is a promising direction for future work. Furthermore, the statistical analysis relies on a simplified scene abstraction. Deviating grasping points observed outside the adhesion region $\mathcal{R}_{\mathrm{adh}}$ reflect the limited expressiveness of a single attachment anchor when describing complex tissue configurations at larger distances. Nevertheless, this simplicity is also a key strength of the proposed representation. By relying on local geometric and mechanical cues, the attachment anchors avoid the complexity specific to surgery and naturally recur across different procedures and anatomical regions. For a given dissection target $D$, the relevant attachment anchor is usually located nearby and characterized by straightforward geometric relationships, which explains why it is applicable in so many different surgical contexts.\par%
Future work may investigate the robustness of attachment anchors under more challenging conditions, including extensive bleeding or fogging.
Furthermore, the general nature of the proposed representation suggests its applicability to other surgical procedures that rely on tissue mobilization and retraction beyond colorectal surgery.
\subsection{Conclusion}
In summary, this study shows that attachment anchors provide a concise and understandable representation of surgical scenes and are an effective intermediate representation for learning-based tissue manipulation. By combining mechanical, visual, and semantic information, attachment anchors reduce uncertainty in predicting grasping points and can be reliably inferred from image data. Experimental results show consistent performance improvements, especially in out-of-distribution settings, such as with unseen procedures, surgeons and workflows from an unseen surgical center. Beyond improving performance, attachment anchors provide transparent insight into how a visual system could interpret grasping-relevant situations, supporting verifiability and human oversight. Together, these findings establish attachment anchors as a meaningful step towards autonomous assistance in robotic surgery.
\section*{REFERENCES}
\bibliographystyle{IEEEtran}
\bibliography{bibliography}
\end{document}